# Strategic Decisions: Survey, Taxonomy, and Future Directions from Artificial Intelligence Perspective

Foundation of A Representation Space for Machine Learning

Caesar Wu*

University of Melbourne caesar.wu@computer.org

Kotagiri Ramamohanarao[1]

Institution of Engineers Australia rkotagiri@gmail.com

Rui Zhang[2]

Tsinghua University rui.zhang@ieee.org

Pascal Bouvry

University of Luxembourg pascal.bouvry@uni.lu

Strategic Decision-Making is always challenging because it is inherently uncertain, ambiguous, risky, and complex. By contrast to tactical and operational decisions, strategic decisions are decisive, pivotal, and often irreversible, which may result in long-term and significant consequences. A strategic decision-making process usually involves many aspects of inquiry, including sensory perception, deliberative thinking, inquiry-based analysis, meta-learning, and constant interaction with the external world. Many unknowns, unpredictabilities, and organic/environmental constraints will shape every aspect of a strategic decision. Traditionally, this task often relies on intuition, reflective thinking, visionary insights, approximate estimates, and practical wisdom. With recent advances in artificial intelligence/machine learning (AI/ML) technologies, we can leverage AI/ML to support strategic decision-making. However, there is still a substantial gap from an AI perspective due to inadequate models, despite the tremendous progress made. We argue that creating a comprehensive taxonomy of decision frames as a representation space is essential for AI because it could offer surprising insights that may be beyond anyone's imaginary boundary today. Strategic decision-making is the art of possibility. This study develops a systematic taxonomy of decision-making frames that consists of 6 bases, 18 categorical, and 54 elementary frames. We use the inquiry method with Bloom's taxonomy approach to formulating the model. We aim to lay out the computational foundation that is possible to capture a comprehensive landscape view of a strategic problem. Compared with many traditional models, this novel taxonomy covers irrational, non-rational and rational frames capable of dealing with certainty, uncertainty, complexity, ambiguity, chaos, and ignorance.

*This work is partially supported by The the Luxembourg National Research Fund (FNR) under grant C21/IS/16221483/CBD

[1] Former Professor of the University of Melbourne

[2] Visting Professor of Tsinghua University and former professor of the University of Melbourne

## 1 INTRODUCTION

Our life is full of choices. Our past decisions make who we are, and our current judgment will make whom we will become. This study primarily focuses on Strategic Decision Making (SDM) for an organization through careful assessments and deliberation. We pay special attention to Strategic Decision (SD) frames. A good SD will make an organization flourish. A bad one will lead an organization to catastrophe. Many SD makers have sought to understand how and why SDs are made in what context [1] [2]. Moreover, they also want to automate Decision-Making (DM) processes by taking advantage of new technologies, such as AI/ML [3] [4] . Historically, the SDM process usually requires tremendous resources and time. However, the results still seem to be arbitrary. The final resolution often depends on the intuition and experiences of select individuals.

With the current advance in AI/ML and other computational technologies, the analytic process of SDM has become much more powerful, foreseeable, reconfigurable, trustworthy, transparent, flexible, scalable, and cost-effective. Although many scholars have made significant progress regarding framing and knowledge representation [5] of ML [6] [7] and AI [8] for complex problem solving [9] [10] [11] [12] in practices [13] [14] [15], there is still a large gap in decision framing and modelling for SDM, which is "a series of associated knowledge representations or logic statements stored in our memory."[16] Most previous studies often focused on rational reasoning for a particular application [17] [18]. However, rationality alone would not be able to solve all our problems, especially for an SD. We often make SDs based on our values, personal beliefs, and psychological emotions or passion. Clausewitz [19] summarized these elements (passion, probabilities, and reasons) in "the paradoxical trinity". Simon [20] defined it as "bounded rationality". Damasio [21] and Rolls [22] illustrated that many decisions primarily depend on our emotions rather than logic or reasoning alone from a neuroscience perspective. Minsky [23] argued that emotions are different ways of thinking. Therefore, this paper intends to conduct a comprehensive survey and articulate a taxonomy of DFs, including rational (Knowledge), irrational (Emotions), and non-rational (Data) decision frames (DFs) [24].

### 1.1 AI/ML In DM Process

The study of DFs for SDM has a fundamental challenge. According to Schoemaker [25], the nature of our contemporary business environment is shifting from certainty to chaos (See Figure 1). Strategic prediction becomes increasingly complex with growing uncertainty, ambiguity, and chaos. Niels Bohr stated, "It is difficult to make predictions, especially about the future.". Schoemaker argued that traditional tools are not enough to manage states of chaos because the world has become much more complex. There are many possibilities and variances in the wide knowledge spectrum, which demand generating multiple DFs, reviewing deep assumptions, and exploring different unknown territories. Schoemaker offered a set of novel solutions in contrast to the traditional toolkit and emphasized "Systems dynamic modelling" to increase our capability to explore, exploit, and test our multiple hypotheses with different DFs. The ultimate goal of developing various DFs from the environment to organic (left to right) and abstraction to reductionism (top to bottom) is to enhance the learning capability and to cope with the shift from certainty to chaos.

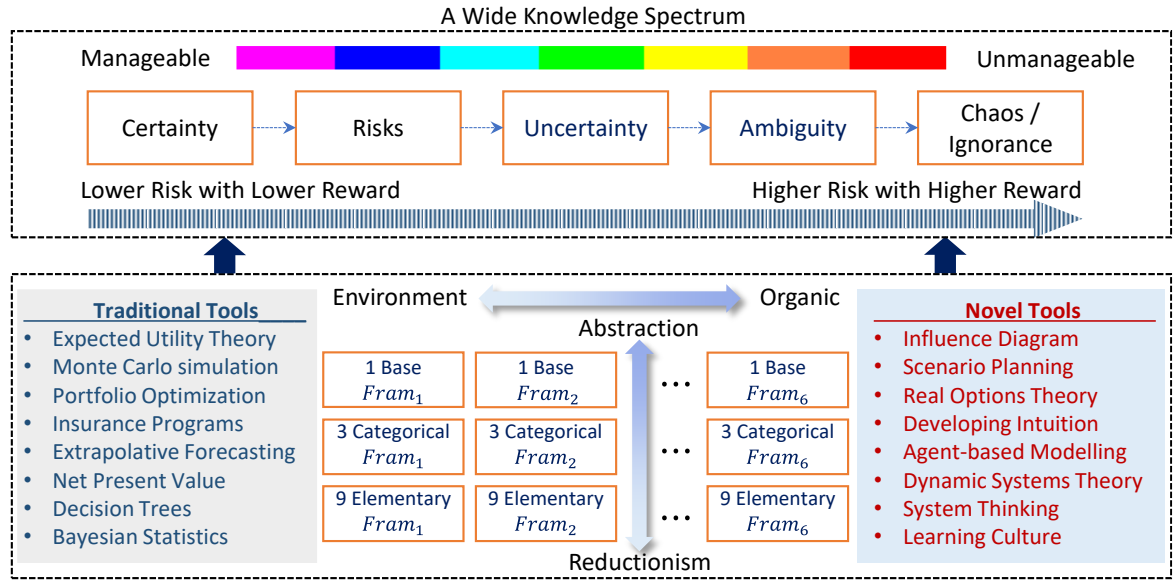

Figure 1: A Wide Knowledge Spectrum Underpinned by DFs with Various Decision Tools (More Details of DFs in Section 4)

A lexical definition of learning capability is a person's ability to comprehend and understand the world and profit from one's experiences by taking multiple DFs, which is an essential part of human intelligence. However, we may become overwhelmed when we face many DFs. Each frame could also have numerous assumptions that are continuously updated due to interactions with the external world. By leveraging AI/ML, we can tell a machine what we want (output or the final strategic goal) rather than what to do (rules) because we often do not know what the optimal solution (decision rules) could reach the final goal. We let the machine find the optimal solution for us. We consider the capability of AI/ML as part of our intelligent faculty for an SDM process. Figure 2 illustrates how to use Decision Framing and AI/ML processes for SDM. In other words, we create multiple frames with different hypotheses to feed AI/ML. We let the machine find a set of the optimal decision rules with a given or desired SDM and dataset rather than given decision rules and datasets for an SD. The logical process is reversed when compared to Good Old-Fashioned Artificial Intelligence (GOFAI).

The essential proposition of adopting AI/ML is that machines will help us review a valuation landscape in the representation space (or model), which is often very challenging to define explicitly. The subsequent issue is how to determine a representation space that is broad enough to include all possible decision rules. On the other hand, it may also be desirable to set the representation space small and precise enough for a machine to learn with less time and resources. There are numerous ways to determine this representation space. With all these goals in mind, we survey comprehensively and propose a two-dimensional decision taxonomy to enable the machine to search for possible rules on our behalf.

## 1.2 Primary Contributions

By exploring and exploiting various DFs in a two-dimensional space, this paper makes the following contributions from an AI/ML perspective:

1. We provide a comprehensive survey of decision frames for various domains, including decision framing bias, corporation planning, AI and robotics.

2. The study presents a novel taxonomy of SDM frames that consist of a total of 54 DFs. These decision frames are derived from rational, irrational and non-rational domains.

3. The taxonomy lays the groundwork for us to deploy five different ML algorithms' tributes (based on their origins) [26] in the learning space or model.

4. The uniqueness of this taxonomy is that it combines Bloom's classification principles [27] with the logic of reductionism and abstract reasoning.

5. In contrast to the previous dichotomy of subjective and objective classification, this study focuses on the organism and its environment of different DFs.

6. This study sets a stepping stone for improving SDM capability by leveraging AI/ML.

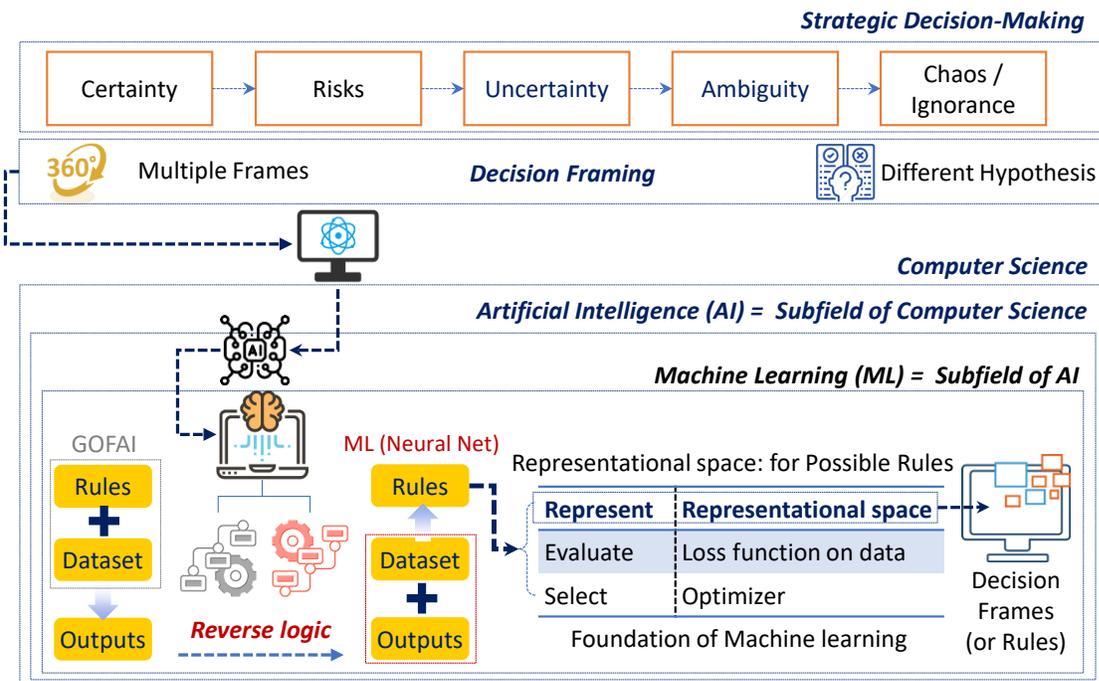

Figure 2: Processes of Decision Framing and AI/ML for SDM

## 1.3 Scope of This Study

The rest of the paper is organized as follows: Section 2 is a simple introduction to the method of how we classify SDM and DFs. Section 3 first articulates the concept of SDM and then classifies some related terms into six groups under the umbrella of what it means to be "strategic". Then, we argue why we want to focus directly on DF instead of SDM. Section 4 provides an extensive survey and details of the novel taxonomy of DFs. The taxonomy consists of three layers based on reductionism logic and abstract thinking. Section 5 discusses the implications of the new taxonomy. Section 6 highlights future challenges, conclusions, and our view of the future direction for SDM research.

## 2 RESEARCH METHOD

We can adopt different research methods to study DF, not limited to statistical, observational, case-study, quantitative, qualitative, experimental or nonexperimental, etc. We classify these methods into four categories: observation, quest, inquiry, and empirical design [28]. Among them, the most compelling category is the inquiry method (See Figure 9 in Appendix) because the characteristic of the survey and taxonomy is exploratory, explanatory, and descriptive. The exploratory survey aims to learn more about a topic than previous researchers have done. The descriptive study aims to answer why something (e.g., a problem) is the way it is. Explanatory classification is to answer why and how questions. In addition, we also combine Bloom's taxonomy [27] approach for various inquiries because Bloom's method is a top-down classification approach that fits our purpose. We intend to figure out how things stand for their principles first and then go from there to infer how the theories can be applied in practice.

## 3 STRATEGIC DECISION-MAKING, CLASSIFICATION, FRAMING, AND INTELLIGENCE

### 3.1 Strategic and Representation Model

SDM is often very challenging to define because "strategic" has almost become anyone's means to an end. We can easily list at least 81 terms that may fall under the umbrella of "strategic". Although " strategy " originates from warfare, it is now part of our everyday vocabulary. Murray [1] indicated that "the concept of "strategy" has proven notoriously difficult to define.", and many theorists failed to clarify the essence of the meaning because "theories all too often aim at fixed values, but in war and strategy, most things are uncertain and variable.".

Although a strategy can be vague, it does not mean it is undefinable. The common definition is that executing a strategy usually has long-term and profound impacts beyond the ordinary and fragmental. It is often contrasted to tactical and operational decisions, which are short-term focused and isolated.

Historically, "strategic" is derived from "strategy". The lexical meaning of strategy is a plan of action designed to achieve an enduring or overall goal rather than isolated objectives. The origin of strategy is drawn from the Greek word "strategia", which stands for generalship. Therefore, it also represents the art of planning and directing overall military operations in a war.

Practically, we can find that many business terms are associated with "strategic". One of the primary terms is "strategic management". Traditionally, strategic management [29] often uses case studies to develop future business strategies. It is compelling for a particular or static environment. However, it does not fit into a complex and dynamic situation because we constantly need to alter our current view and update our representation model in our memory or a database. Minsky called the model a frame that is "a data structure for representing a stereotyped situation…," "We can think of a frame as a network of nodes and relations.". [30]

### 3.2 Why Decision Frames

Minsky's definition gives us some clues on how to create "a data structure", which simplifies and summarises a large quantity of information so that a user can make sense of it. Consequently, we must pay attention to some information and ignore others. However, different ways of framing will lead to different decisions. For example, we can frame 90% success of a business strategy, and we can also frame 10% failure. The only difference is that we use different lenses or look from different perspectives. Numerically, it should not matter how we frame the same information, but practically, framing can influence our DM process profoundly. If we frame "success," people will shift their attention to success. If we frame "failure," people will worry about risks. Therefore, framing does matter. More importantly, DF is oriented by decision rules rather than the final result, which fits the criteria of the ML process (See Figure 2).

Russo and Schoemaker [31] proposed a four-stage DF process that consists of Stage 1: framing; Stage 2: gathering intelligence; Stage 3: coming to conclusions; and Stage 4: learning from experiences (see Figure 3). They argued that the process is best practice because it can prevent many characteristic errors and common decision traps. Although the process does not have a series of rigid rules, it is a framework that can be applied practically. However, the cumulative experience-based process does not fit with AI/ML. We need DFs from an AI perspective.

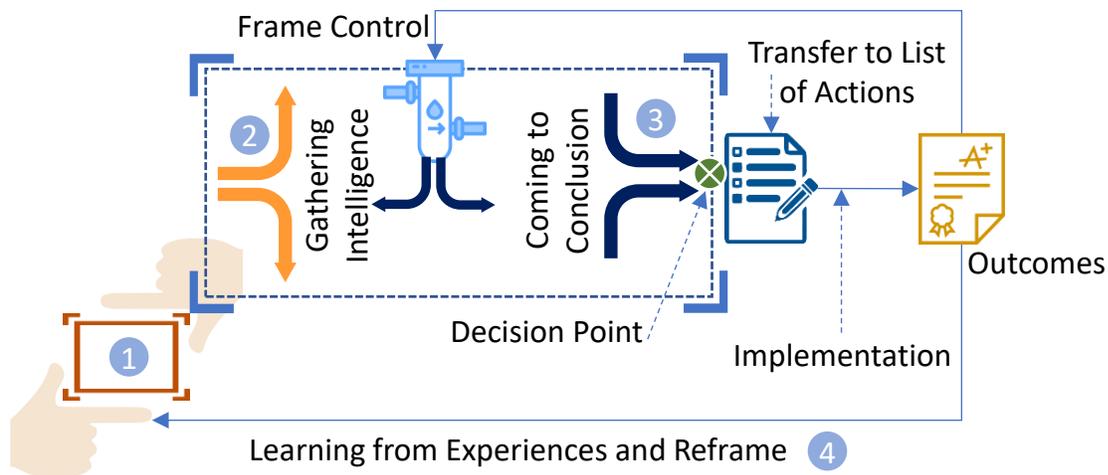

Figure 3: Decision Framing: A Good Decision-Making Process

Our goals for creating comprehensive DFs are 1.) To learn interdisciplinary decision rules, 2.) To cope with an increasingly dynamic environment, 3.) To communicate our ideas with others effectively, 4.) To reverse the logic of AI/ML, 5.) To leverage AI/ML capability, 6.) To solve issues that require cumulative experience and training. Practically, we can develop as many DFs as we want. However, the question is whether we can organize these DFs as a coherent structure that allows a machine to learn with reasonable computational time. It is a very challenging issue. Therefore, we first review some selected seminal papers to see how previous scholars respond to this question.

## 4 SURVEY AND DEVELOP NEW TAXONOMY FOR SDM

### 4.1 Survey of Decision Frames

Carter et al. [32] adopted the combination of qualitative cluster analysis with Q-sort methodology to develop DM biases from a human behavioural perspective for the supply chain management. They articulated 9 DM bias clusters based on 76 decision biases and argued that these biases could be generalized for DFs. However, they also noticed that their proposed taxonomy requires further testing because it is just a theoretical proposal. Nutt [24], on the other hand, analyzed 352 strategic decisions and unveiled different factors that could influence DFs. These DFs could lead to different strategic directions. The author intended to highlight the best practice for SDM.

De Jaegher [33] addressed strategic framing from the perspective of behavioural economics or prospect theory. Although the paper discusses prospect theory, the underlying DF construct is like a game theory construct between patients and physicians. The issue between patients and their physicians is less strategic than tactical because most cases are less ambiguous and significant. Arend [34] proposed a three-step process for SDM to overcome the issue of ambiguity. At the core of Arend's approach is an ex-post backward inductive logic reasoning plus conditional probability. The terms can be different, but the essential DF that the author adopts is based on game theory and expected utility payoff. Haksever et al. [35] defined a value creation type of DF to improve key stakeholder relationships. The authors outlined five different scenarios that could impact a strategic DF. Schoemaker is a pioneer of scenario planning methodologies [36] for SDM (Refer to Fig.1). Schoemaker [37] proposes a strategic radar or decision framing apparatus (three elements) with five stages to integrate weak and emerging signals in a chaotic and noisy environment.

Another strategic management pioneer, Henry Mintzberg [38], argues that we should clarify the strategic concept from different strategic perspectives. Mintzberg provided five practical ways to frame an SD, known as the 5Ps or Planning, Pattern, Position, Perspective, and Ploy. In addition, Mintzberg also proposes a handy tool for strategic framing known as the ten schools of thought [39]. These schools of thought can be divided into two groups (three schools for looking forward and seven for reasoning backwards). Although these narrative tools are convenient and practical, they are unquantifiable from a computational perspective. Nevertheless, these are excellent ideas that have laid the groundwork for further SDM research.

Bateman and Zeithaml [40] demonstrated the ANOVA or empirical method to find an SD emerged from a stream of incremental decisions. The DF is derived from the psychological context, including the past, present, and future outlooks as three experimental variables. From an AI/ML perspective, Minsky contributed considerably to psychological decision framing during the 1980s and 2000s. His landmark books "Society of Mind" [41] and "The Emotion Machine" [23] illustrated how to develop a mind-frame for knowledge representation [30]. A critical aspect of Minsky's work is that he deliberately blurs the line between computer science and psychology for the next generation of researchers.

By contrast, Bouton, Maxime et al.[42] provides a computational solution for framing a decision problem, known as Partial Observation Markov Decision Process (POMDP), to simulate the urban mobility of autonomous vehicles for an uncertainty problem of autonomously navigating urban intersections. The authors use the transition and the observation models to represent the DF and implement it with the online algorithm and the interacting multiple-model (IMM) filter. The paper demonstrates that its empirical result is better than a threshold-based heuristic decision strategy regarding vehicle safety and efficiency. However, as the authors indicate, the result would be more convincing if their solution could include pedestrians or other variables in the DF model.

Siagian et al.[43] provided a solution for an autonomous mobile robot navigating a pedestrian environment. The system known as Beobot 2.0 can effectively walk through a large crowd. The paper argued that the success of their solution was due to hierarchical map representation primarily guided by the vision. The essence of their DF is the new hybrid topological, which is a hierarchical spatial representation (a global topological map) with a local grid-occupancy map. However, the success rate is dependent on the travel distance. The longer the travel distance, the higher the failure rate.

In summary, each DF has its pros and cons. Table 1 outlines each method, its primary contributions, advantages and potential gaps. However, the critical issue remains: how can we use various DFs to establish a coherent framework for ML?

We argue that the taxonomy of DF is an essential part of scientific inquiry for SDM because it can help us cope with the complex, chaotic, and uncertain world [44]. It can create coherent principles for us to recognize an object quickly and efficiently [45]. Scherpereel [46] proposed a decision taxonomy based on "semantic descriptors" or by performing a content analysis on the seminal literature in the natural, social and applied sciences. Scherpereel called it a "decision-order", which is a hierarchical structure that includes three layers (1st order, 2nd order and 3rd order).

### 4.2 The New Taxonomy

Based on the previous studies, we consider a taxonomy of DF as a typology of various mental models [47] to understand the complex world around us from different perspectives. Hernes [48] suggested that organizations should be as processes "in the making" rather than "things made.". It means the DF should focus on the relationship between organism and environment.

Minsky [30] also suggested that each frame should be associated with different kinds of information. Some information is about what will be expected next, and others are about what to do if the expectations are not met. Minsky argued that framing only represents a "problem space" for heuristic searching, but it is not well defined for it to be useful to a computer programmer. Therefore, we create various frames from three perspectives: logically, empirically or interactively (in a spatial and temporal domain), and psychologically.

Logically, Hayes [16] suggests that the notation of frame is nothing but a series of associated knowledge representations or logic statements stored in our memory, "which can be retrieved via some kind of indexing mechanism on their names." Baron [49] explains that the meaning of logic is in the theory of thinking. A logic relation only means inference but not understanding errors resulting from this inference. In common sense, logic represents "reasonable" or "rational." However, there are many ways to be "reasonable". From a formal perspective, Baron exhibited three kinds of logical systems, namely, a.) propositional logic (e.g., a relationship of "if-then"), b.) categorical logic (e.g., a relationship of "all, some, none, not, and not") c.) predicate logic (e.g., specified a relationship between two terms). The formal logic reasoning systems provide absolute certainty for a conclusion. However, it does not care whether the premises are true or not. Moreover, the formal logic does not provide us with new information.

To improve the formal logic reasoning, Johnson-Larid [47] put forward the mental model proposition, which is to build a mental model to reach a possible conclusion for a particular circumstance and assumptions. Thus, we can derive logical content and context frames.

Empirically or interactively, Horn [50] suggested that framing concerns a building block of communication. The frame signifies how to interpret and classify the obtained information to the audience. The implication of framing is how to encode the real meaning into a message so that receivers can decode the message regarding its relationship with their existing beliefs. Framing sparks the meaning from the receivers' perspective. Horn argued that framing could also be interpreted as integrating old with new information in the temporal and spatial domains. Consequently, two frames are critical when we interact with the world: time and space.

Psychologically, Amos Tversky and Daniel Kahneman [51] defined "frame" as a mental model to characterize possible outcomes from risky perspectives regarding values loss and gain. Ultimately, a decision quality depends on value creation [52]. Keeney [53] illustrated how to create values for decision alternatives. Keeney demonstrated a reactive approach to framing a set of decision-making alternatives and creating potential values for an organization. In short, a mental model is related to our values and emotion.

In a nutshell, decision framing means different presentations and interpretations of a decision problem. It influences us to make different choices, although the underlying decision problem could remain. Based on three perspectives of framing, we can heuristically generate six basic categories known as "content", "context", "spatial", "temporal", "value", and "emotion". The reason for identifying these six categories is that they include rational (reasons or knowledge), non-rational (probabilities or data), and irrational (passion or emotions) minds of human thoughts. The interpretation of these frames depends on information received, knowledge held, assumptions made, and even embedded emotions. The world is a multidimensional object. We can focus on different parts of the object by adopting different frames. Therefore, we have chosen 6 base frames at the top layer, 18 categorical frames in the next layer and 54 elementary frames at the bottom. The classification aligns with three levels of DM, namely strategic, tactical and operational (See Figure 4).

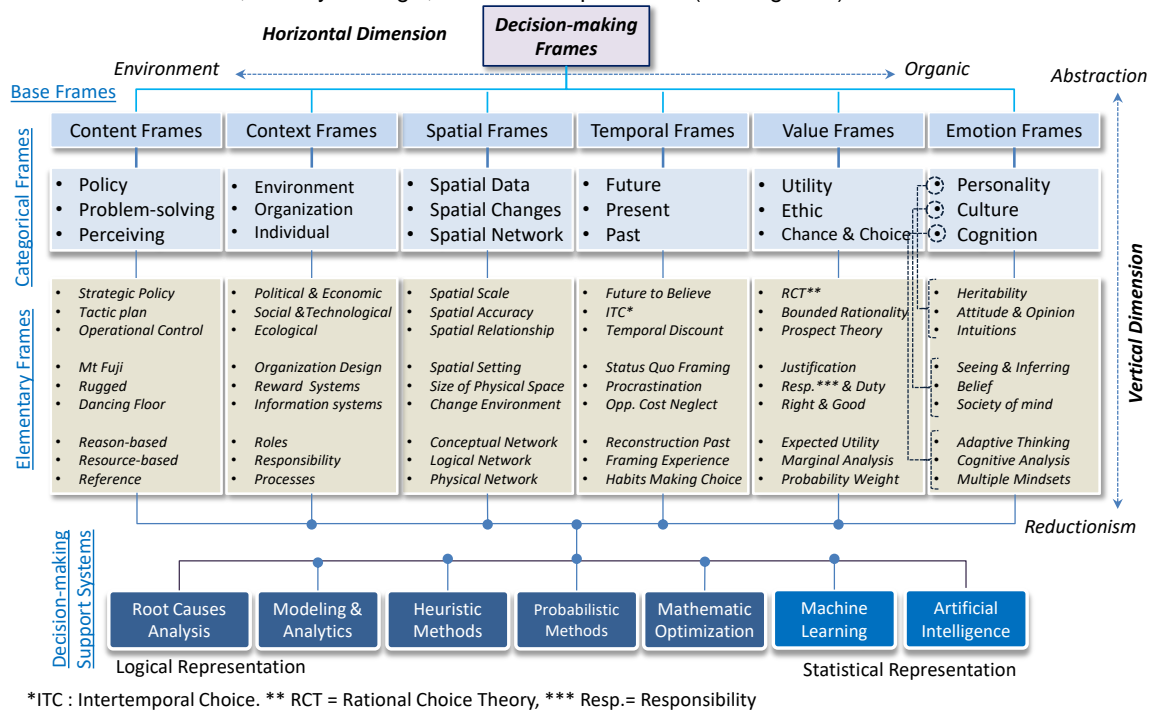

Figure 4: The Taxonomy of Decision-Making Frames with Various Decision-Making Apparatus

Supporting these 54 frames, we have seven apparatuses as decision support systems (Root Causes Analysis, Modeling and Analytics, Heuristic Methods, Probabilistic Methods, Mathematical Optimization,

Machine Learning, and AI) to analyze a given decision problem in a particular problem space. Thus, a taxonomy has emerged as the novel framework that consists of three layers that can shape different SDM problems shown in Figure 4. We now look closely at each base and categorical frame in detail. [3]

### 4.2.1 Emotion Frames

Among many irrational frames, our emotion is essential to our judgments about right or wrong. Emotion provides the foundation for our moral capacity. Traditionally, emotion is considered a distraction rather than a value-added element for DM because it is irrational. Aristotle explained that emotion is "a more or less intelligent way of conceiving a certain situation, dominated by a desire." [54] Although there could be different emotions, they "are those things by the alteration of which men differ with regard to those judgements which pain and pleasure accompany, such as anger, pity, fear and all other such and their opposites." [55] William James believed "emotion is just a physiological reaction, essentially, its familiar sensory accompaniment – a 'feeling' ."[56]

Emotions refer to various experiences that involve some appraisal of inner thoughts, feelings, memories, motivations, and bodily reactions.[57] Today, a simple definition of emotion is an integrated and adaptive response of mind and body to a stimulus of rewards and punishers. They support our survival. Emotions have three primary purposes: 1.) they notify us about important events, both good and bad. They get our attention to focus on some crucial event; 2.) they motivate us to behave in ways that deal with an event, 3.) they generate changes in our body. Many of these changes are designed to prepare us to respond to whatever caused the emotion. According to the neuroscientist Damasio [21], people cannot make a decision if the emotion is absent because people will keep reasoning and deliberating instead of committing to an action. Damasio called it a "somatic marker hypothesis" or "as-if" loop. To understand how our emotions impact our DM, we develop three categorical frames: personality, culture, and neuroscience. We use personality to study individual influence, culture to investigate social manipulation, and neuroscience to analyze the human brain's impact.

### 4.2.1.1 Personality

Personality refers to an individual's characteristic pattern of thinking, feeling, and acting [58]. It answers the question of why each person turns out differently in terms of DM. Generally, personality is often a relatively complicated question and ends up with a complicated answer. Traditionally, we consider personality neither pure nature (what we were born with) nor absolute nurture (how we were raised). Studies [59] show both nature and nurture effects involving genetic and environmental influences for a personality to be cast out. Under the personality frame, we formulate three elementary frames to understand people's DFs. These are the heritability coefficient, attitudes, and intuitions.

Heritability refers to the proportion of the observed variability in a group of individuals that can be accounted for by genetic factors. A percentage of phenotypic variance (we can observe in a trait or characteristic, such as music talent) is due to genotypic variance (changing rate in people's genes). Most of the heritability coefficient observed is between 0.2 and 0.5 across people.[60]

---

[3] Notice that the explanations of different frames could be intertwined. For example, we use the concept of "belief" to explain the ethical frame and use bounded rationality to define the future. It is just the nature of the human concept. It is similar to the definition of colour and yellow circulating, in which seven primary colours also define colour. Yellow is one of them. When we describe yellow, we explain it as one of the primary colours.

These genetic differences will contribute to a person's attitude (personal feeling or opinion about something) and value. One example is the attitude towards either liberal or conservative opinions, e.g., tolerance, openness to new ideas, stubbornness, practicalness, etc. To a certain extent, an attitude is a convenient shortcut to making a quick decision so that people can respond to a particular thing immediately.

Likewise, we can also consider intuition a convenient mental shortcut, but it is a pattern match mechanism (via unconscious mind or gut feeling), which contrasts with decision deliberation (via conscious mind or logic reasoning). If "attitude" emphasizes a person's action towards someone or something, then "intuition" focuses on a person's competence through a long time of training.

In principle, we can adopt three elementary frames to understand SDM when time is limited; heritability, attitude and intuition. Heritability shapes the genetic differences that may impact a person's decision. Attitude identifies a person's readiness to respond to all situations, while intuition defines a person's decision competence due to long-term training and learning. Whether it is heritability, attitude, or intuition, the cultural environment is one of the primary factors determining these elementary frames.

### 4.2.1.2 Culture

Benet-Martinez argued, "Culture is a key determinant of what it means to be a person."[60] Culture is something we take for granted about how things work around us, and it outlines how we approach a decision problem. Bucholz [61] highlighted, "geography is destiny; culture is reality.". Culture impacts the way we frame and deal with a problem. We see culture as a set of glasses that could shape our opinion or way of seeing the world around us. If we consider culture as an allegory of various glasses, three elementary frames are derived: seeing and inferring, belief, and society of mind.[41]

An example of cultural difference is in mathematical thoughts. In the east, people think of mathematics as an application tool, which began with various practical problems, including land surveying, building a canal, calculating tax rates, and predicting the sun's position for agriculture activities. In contrast, the West thinks of mathematics as the heart of education. Plato's famous quote: "Let no one ignorant of geometry enter here." Eastern culture sees a reciprocal relationship of opposite forces (e.g., Ying and yang or Taichi), emphasizing collective efforts. Eastern culture infers an overall harmonious partnership. Conversely, the Western culture sees the arch-rival of the opposition and emphasizes arguments in the debate to prove the truth and shoot down the false.

Our cultural background will influence personal beliefs when information is insufficient to make an SD. Our beliefs will influence our decision and actions. Nilsson [62] described it as "a proposition that one holds with a strength that could be very weak, very strong, or anything in-between." Jervis proposed, "The concept of beliefs has several connotations, some of which involve faith and emotions." [63] To quantify belief, we can approximately draw a belief spectrum between 0% and 100% of evidence between faith and knowledge (see Figure 5).

On the superficial level, faith seems redundant in terms of belief. However, faith implies "belief without sufficient evidence or justification." To the extreme, blind faith stands for believing without reason, which is turning a blind eye to any evidence or irrational. By comparison, knowledge is a justified true belief (or rational). Technically, the "p" value is less than 5% (unjustifiable reason). "Conditional belief" is anything between faith and knowledge, in which an initial belief probability is waiting for new information to be updated (or non-rational). The question is, where does human belief come from? It comes from mental constructions through other beliefs, learning processes, and experiences [62]. According to Minsky's theory, the working process of belief is through the society of mind [41] (many mindless agents) to formulate mental constructions.

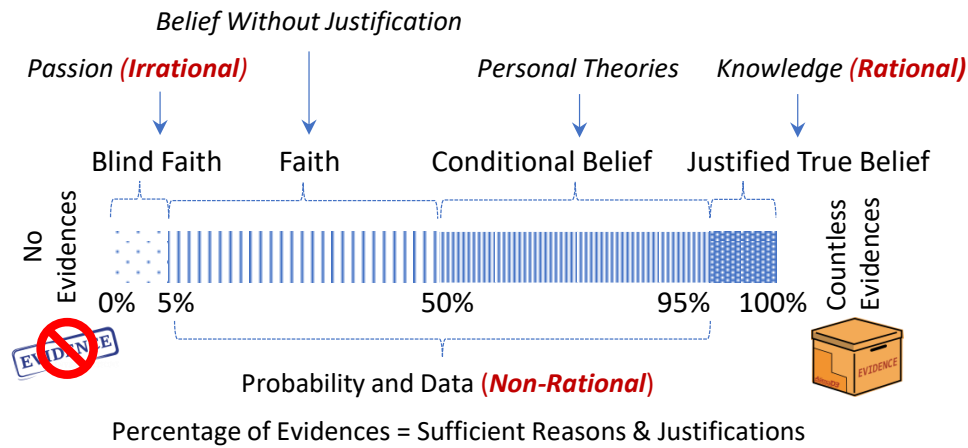

Figure 5: Belief Spectrum

From a cultural perspective, we can summarize three elementary frames to underpin the DM process: seeing and inferring belief and society of mind. Minsky's core idea of how the mind works is that "minds are what brains do." This proposition gives rise to the framing of cognition.

### 4.2.1.3   Cognition

Cognition refers to an information process via various mental activities or thinking. Cognition aims to obtain comprehension and knowledge. The origin of the word is derived from the Latin "cognitio", which came from "cognoscere". It consists of "co" (intensive) and "noscere" (to learn). Recently, many studies focused on how people make a "good enough" decision in their thinking process, known as cognitive psychology. The theme is to study thinking about thinking or a meta-thinking process.

To frame from a cognitive perspective implies leveraging cognitive psychology to improve our SDM. We derive three elementary frames: adaptive thinking, cognitive analysis, and multiple mindsets. These frames improve our cognitive capability when we encounter complex problems. Human does not have unlimited cognitive capacity and time to gather all the information and deliberate all the possible options for a particular decision. We usually work out our solution with many mental shortcuts.

The adaptive thinking frame is to be open-minded or open to new information. It is also dependent on responsiveness to feedback. Part of the responsiveness enables the DM process to explore further information from multiple sources associated with present and possible future conditions. The frame can justify habitual thinking patterns along with environmental changes continuously and spontaneously

Our cognitive limitations and time constraints forbid us from handling much data for the DM process, even with powerful modern computers. As Heuer pointed out [64], "People construct their own version (frame) of 'reality' on the basis of information provided by the senses, but this sensory input is mediated by complex mental processes that determine which information is attended to, how it is organized, and the meaning attributed to it." Subsequently, we may seek our own version of reality and satisfaction. It also means bounded rationality.

Bounded rationality theory argues that human decision is not optimized to the way traditional economic theory presumes but to human satisfaction. By satisfaction, we explore possible alternatives only to the point where we find a reasonable solution that is satisfied with ourselves. We do not keep looking for the ideal solution. Very often, we use a bypass. These detours introduce many cognitive biases, such as over-confidence, sunk-cost, recency effect, etc. Moreover, our decisions can be easily influenced by the presented options or "framing-effect" [65], but other emotional frames may mediate it to reach approximated.

In addition to cognitive analysis, multiple mindsets also help us with SDM because our minds often trick us with easy and quick solutions to deal with the complicated world. These quick responses could form a mindset quickly and resist change because we think we already know. Consequently, we only see what we want to see and ignore what we do not want to see. Multiple mindsets frame can help us overcome the issue of a single mindset. In addition to emotional frames, we must also understand value frames because value permeates our life.

### 4.2.2  Value Frames

We cannot see the world without value lenses because "value" formulates our idea of "what is worth living" and "what is worth striving for". Economically, the core issue of SDM is one of value judgement. If we only have facts without value, we cannot decide because all facts have the same status. Without a doubt, science deals with a body of facts systematically and can help us understand all the facts. Science puts all facts into a complete picture of the real world. However, it is still not enough for us to decide because "what we should do about the facts" is above and beyond questions of "what has happened", "what will happen", and "what-if." The issue of "what we must do" is the question of values [66]. If we ponder the question of values, we should clarify utility, ethics or morals, plus chance and choice frames because they underpin our values. These notations explain "what is good to be good?" and "what is right to be right?" [67].

#### 4.2.2.1  Utility

The concept of utility has multiple connotations. The original definition of utility can be traced back to Jeremy Bentham's "Introduction to the Principles of Morals and Legislation" (1789). Bentham argued that "utility is meant that property in any object, whereby it tends to produce benefit, advantage, pleasure, good, or happiness…." Bentham's principle of utility (Good-Bad theory) for the good decision is to produce "the greatest amount of pleasure for the greatest number of people." [68] Bentham's utility theory leads to a core economic concept: rationality. Based on this core idea, Simon [69] questioned the rational choice theory (RCT) and established the bounded rationality theory [20], [70] for the DM process. In 1979, Kahneman and Tversky [71] proposed the prospect theory for DM under uncertainty. Based on these thoughts, we can develop three elementary frames: rational choice theory, bounded rationality, and prospect theory for the utility frame.

Although Bentham's good-based theory is quite powerful, it cannot be absolutely correct because it has no respect for the basic value of anything (e.g., human life and human dignity). Therefore, we must introduce an ethics and morals frame or right-based theory to solve the issue.

#### 4.2.2.2  Ethics and Moral

Ethics (Right and good theory) is "the study of the choices people make regarding right and wrong."[72] It investigates a set of principles or accepted codes of conduct that define how we should behave when acting in a public place. The word "ethic" is derived from the Greek "ethickos", which means ethos. It implies the fundamental character of the spirit of a culture. It is a set of beliefs and customs about a person or group's social behaviour and relationships. Often, ethics set a higher (or maximum) standard, i.e. honesty, honour, integrity, and excellence, in practice for professionals, including accountants, teachers, journalists, physicians, senior business executives, and lawyers. In contrast, a law is a minimal standard of ethics that can be codified and enforced. The law regulates what we should not do, and the ethical codes define what we should do.

Theoretically, Graham [73] highlighted eight moral theories: egoism, hedonism, naturalism, existentialism, Kantianism, utilitarianism, contractualism, and religion, for decision-makers to consider how to make their ethical and moral judgments. However, many ethical and moral theories may conflict with each other in the ethical DM process. Schwartz [74] proposed a unified approach to address the issue. The approach includes a "person-situation" interactionist process and an "intuition/sentimentalist-rationalist" method for moral judgment. Similarly, Tenbrunsel and Smith Crowe [75] summarised ethical and moral decisions into moral awareness and moral and amoral DM. In addition, Audi proposed ethical intuitionism as an alternative to Kantian ethics and Benthamism utilitarianism [67]. Altogether, we should have three elementary frames known as justification (integrated both ethics and morals), responsibility and duty (a combination of Benthamism and Kantian ethical theories), and right and good for the value-based DF.

However, Singer [76] argued that despite many ethical theories that have been developed, there are still some uncertainties about what exactly we should do and justify what we are doing when we make ethical and moral decisions. The issue of uncertainty results in framing chance and choice.

### 4.2.2.3 Chance and Choice

The utility framing assumes certainty, but nature is uncertain and constantly changing. We must make a series of interactive decisions to cope with the dynamic world. Uncertainty is a fundamental and escapable part of our life. When we bring utility and uncertainty together, we have one of the economic core ideas - the expected utility theory. Under uncertainty assumption, it explains an agent's optimal choice to gain the highest expected utility for an SD. The expected utility determines the weighted average of all possible utility outcomes [77]. To weigh all possible utility outcomes [78], we also need to estimate a frame's margin. This estimation brings out the idea of the value of trade-offs. Thus, we can derive three additional elementary frames defined from chance and choice: expected utility, marginal analysis, and probability weight. All values have temporary stamps (e.g. net present value) by introducing uncertainty. Therefore, we derive temporal constraints from the uncertainty principle for DM processes.

### 4.2.3 Temporal Frames

Temporal framing fabricates possible decision options for the future. We inherit the past, inhabit the present, and image the future. Our decision over our entire lifetime is similar to a tree. The past is like a single branch or a single line. Many previous options had already been closed to us. We cannot alter the decisions we have already made, but we can review all past decisions. Our past choices lead us to our current situation. Many future possibilities are dependent on a series of our current decisions. When we are young, we have many alternative paths. When we are old, we only have fewer options in the future.

### 4.2.3.1 Future

Human has a unique capability of imagining the future. We might also call it scenario planning. According to Russo and Schoemaker [31], the scenario is a "possible future." These future scenarios are narrative stories that serve our thinking and stretch our present frames or imagination. It is one of the new tools in the DM process (See Figure 1). Scenario planning aims to mitigate future uncertainties within certain boundaries for SDM. Scenario planning offers a valuable function for a reasonable assessment. Scenario planning is simulation and emulation from a computer science perspective.

Strategic decision-makers always try to broaden their horizon of knowledge because the sooner they know, the better they will prepare. The more they know, the more options they have. However, we have often been immersed in a vast amount of information and have very little time to react. We want to push out our future frame as far as possible, but the future is approaching us simultaneously. The future presents us with a paradox. We propose three elementary DFs to deal with the paradox: "future to believe", "Intertemporal choices" (ITC), and "temporal discount" to evaluate the possible future impacts.

Uncertainties suggest both risks and opportunities. The most crucial question for SDM is "what is the future to believe?" when facing a competitive environment. We often believe more about what we know than what we unknow. The framing of "future to believe" implies knowing ourselves (resource, capacity, weakness, and ability), our competitors, our environment, and our relationship with others.

The "future to believe" help decision-makers to create a decision strategy of intertemporal choices [79], which is what to do at different strategic decision points in a temporal dimension. It means trade-off risks with opportunities or costs with benefits at different points in time. [80]. The frame of intertemporal choices leads to the idea of temporal discount, which is a reduction in value because of the expected passing time. Different people would have different temporal discount functions. The reasons for temporal discounting are 1.) risks of the future rewards, 2.) temporal discounting due to temptation of other things, and 3.) we prefer our present self to our future self. These reasons give rise to consideration of present framing for future rewards.

### 4.2.3.2 Present

We know that different frames can emphasize various aspects of a decision problem. Consequently, we can make one aspect of a decision problem more important than others. Although the contents of a problem remain, we change the frame of the decision problem. There are multiple ways to frame the present situation. When the future arrives, we often prefer keeping the status quo, which is doing nothing and upholding previous decisions. Samuelson and Zeckauser [81] argued that this preference for sticking to the status quo is disproportionally large and has become a dominant factor in people's DM process. Similarly, Johnson and Goldstein's [82] demonstrated a result in their experiment regarding people's decision on organ donations and suggested that adopting a status quo frame would make a substantial difference.

One reason to maintain the status quo could be decisional procrastination, which is one of the DM styles. According to Burka and Yuen [83], the emotional roots of procrastination are triggered by "inner feelings, fears, hopes, memories, dreams, doubts, and pressures. But many procrastinators do not recognize all that is going on under the surface, because they use procrastination to avoid uncomfortable feelings.".

The opportunity cost framing would overcome the status-quo issue. One of the fundamental principles of economics is that there is no free lunch because our demands and limited resources are unavoidably imbalanced. If we decide to divert our limited time or resources to do one thing, then an opportunity for doing another thing will be forever gone. This giving-up opportunity is known as opportunity cost, even if we decide to do nothing. Frederick et al. [84] suggested that bringing the frame of opportunity cost to our mind can substantially shift a decision maker's preference.

### 4.2.3.3 Past

If we shift our reference point to the past, we reconstruct past frames with different attributes and information about the current situation. Ibn Khalldun once said, "The past resembles the future more closely than one drop of water resembles another."[85] It exhibits that we are always trying to make sense of our past and drive the consistency of our beliefs and behaviours.

When we overhaul the past, we inevitably bring negative and positive preconceptions from the past. In comparison with material goods, personal experiences are difficult to define. Still, based on common sense, we interpret them as memories, perceptions, and emotions, which are bought by some previous observations or events. According to Tversky and Kahneman's prospect theory [86], two small gains or losses will be felt larger than a single significant gain or loss because prospect theory argues the diminishing sensitivity for a value function.

If some specific experiences recurred, the repeated experiences would become a personal habit. Verplanken and Aarts [87] defined habit as "… learned sequences of acts that have become automatic responses to specific cues, and are functional in obtaining certain goals or end states." Ouellette and Wood [88] suggested that habits are behavioural tendencies. "With repetition and practice of a skill in a given setting, the cognitive processing that initiates and controls the response becomes automatic and can be performed quickly in parallel with other activities and with the allocation of minimal focal attention." The well-known theory that explains such phenomena is the dual-system model of cognition. According to Kahneman [89], we have both a fast-thinking system (system-1) and a slow-thinking system (system-2). When system-1 is in charge, our reaction is fast, almost spontaneous, automatic, and effortless. In contrast, system-2 is slow, prudent, deliberate, and effortful. This dual-system thinking model illustrates the principle of human habit. We often consider the fast thinking system as trained habits of responding to external stimuli and taking actions only when the slow thinking system approves or neglects the system-1 requests. Therefore, a habitual choice (based on past experiences) merges processes of intuition, heuristics, and automatic decision rules.

### 4.2.4 Spatial Frames

Spatial DFs are often designed to solve complex problems in conjunction with temporal frames. For any complex issue, space does matter. It is self-evident. However, many people often exclude spatial frames when building DFs, such as economical pricing, cultural influence, cognition biases, etc. We often believe that we can decide anything at the head of a pin. There are two reasons for this belief: 1.) We prefer a simplified model. 2.) Spatial framing is often too hard to be constructed. With the advance of powerful hardware and the growing sophisticated software, analyzing spatial frames becomes possible.

The simple definition of spatial is when its characteristics and attributes become the dominant factor in the SDM process. For example, if you decide to buy residential property (as a strategic investment), Harold Samuel's advice is "Location, location, and location." [90] Another example is the story of 'The Ghost Map," written by Steven Johnson [91], in which John Snow used spatial information to identify the contaminated water well as a root cause of spreading the cholera epidemic in 1854 London. Our decisions are also influenced by others due to our social relationships or a virtual network. Spatial framing is often a central theme for a complex system. Lastly, if we alter the size of the physical space, people could make different choices [92]. The spatial DF consists of three categorical frames: 1.) Spatial data, 2.) Spatial changes, 3.) Spatial network.

### 4.2.4.1 Spatial Data

Principally, spatial data studies the locational and distributional arrangement of people, events, objects, and interconnections in space. Processing spatial data aims to uncover underlying patterns and behaviours due to complex interactions among people, events, and objects. These patterns and behaviours are critical for SDM. Spatial data analysis is essential for many businesses, such as airlines, railways, shipping, retail stores, real estate agents, taxis, fast food services, logistics and distribution services, and even fishing and agricultural businesses.

When we pay close attention to spatial data analysis, we have three types of spatial data frames concerned by decision-makers, namely 1.) the scale of spatial data, 2.) the accuracy of spatial data, and 3.) the relationships of spatial characteristics. We can use vectors (e.g., point, line, and polygons), rasters (e.g., thematic, spectral, and imagery), or spatial functions (e.g., comparison, reference, and special defined). Spatial data accuracy is associated with the complexity of a targeted problem within the capacity of computational power. It is an evaluation process to find the desired scope within the given spatial data. The approach could be an iterative process by a trial-error method to find the optimal point. The relationships of spatial characteristics deal with the degree of dependence of spatial parameters. The spatial relationship is often based on Tobler's first law (TFL) [93] "I invoke the first law of geography: everything is related to everything else, but near things are more related than distant things." TFL is the core concept of spatial analysis and modelling.

### 4.2.4.2 Spatial Changes

Spatial changes are physical space variations that may influence people's DM processes and psychological behaviours. Often, we hardly notice spatial variations. Many spatial influences are due to distance, visibility, personal preferences, and locational convenience changes. People are hardwired by shortcuts, rough estimation, heuristic thinking, and intuitive inference. According to Kahneman [89], we intend to use system-1 to make a quick judgement rather than make system-2 (our frontal cortex) because we could overload system-2 if we always use it for every decision. Consequently, we are more likely to capture an object or thing within immediate reach, easier to comprehend, and highly visible to us. In other words, people make different choices with alterations in the architecture of choices. Thaler called it a "nudge"[94]. He showed that people would change their minds if some special arrangements of choices were made without abandoning any options.

Naruse et al. [95] also showed that the previous decisions would usually not impact future decisions if the size of the local environment or space is large enough based on their local reservoir model. On the other hand, if local space size decreases, the previous decisions will influence future decisions. The implication of Naruse's experiment shows that if we want people to change their minds, we should give them limited space. Conversely, we should allow enough space if we want them to stick their minds. In a nutshell, we influence people's decisions by manipulating spatial frames 1.) change the spatial setting (or spatial nudge), 2.) change physical space, 3.) change the environment in terms of colour, style, feature, and illumination to stimulate people's either system-1 or system-2 thinking.

### 4.2.4.3 Spatial Network

The spatial network is another vital factor which influences decision-making. It consists of both edges and nodes. We should observe examples of networks everywhere, such as telecommunication networks, the internet, transportation networks, power grids, ecological networks, and neural networks. There are many different approaches to classifying spatial networks. Yang and Shekhar [96] classify a spatial network into three levels: conceptual, logical, and physical that is from an architectural perspective. Barthelemy [97] defines a spatial network into five categories that consist of geometric graphs (representation), spatial generalization of the Erdos-Renyi graph (driven by probability), small-world phenomenon (a median chain length of a social network is six) [98], spatial growth models (growing process), and optimal networks. Barthelemy's classification is based on spatial network modelling. Oliver [99] categorizes a spatial network based on a data genre. It is the application-oriented approach. From an SDM perspective, we argue that the architecture approach to classifying spatial networks is compelling because the method is possible to handle the spatial scaling with the spatial data. Consequently, we should have three types of elementary frames of spatial network: conceptual, logical, and physical.

### 4.2.5 Context Frames

In addition to spatial frames, Bucur et al. [100] suggest an ambient intelligent approach to develop a context-based decision-making process. To better understand the context-based DFs, we first review the concept of context-free. According to Voors et al. [101], a context-free decision means that behaviours (or DM) can be disconnected from circumstances or facts surrounding a particular event. They raise doubt that any decision is separated from an environmental context. In contrast to context-free, we have context-sensitive [102]. Duffy [103] defines "context" as "circumstances relevant to something" or "the interrelated conditions in which something exists or occurs." Duffy suggests a simple model of context that can assist business executives in establishing context frames for SDM. The frame has three variables: environmental, organisational, and individual. Establishing such context frames aims to help decision-makers cope with uncertainties and complex and chaotic situations (Refer to Figure 1) [104].

### 4.2.5.1 Environment Context

When considering the environmental context, it is essential to include political, legal, economic, social, technological, and ecological variables because they are a part of the complexity of a decision environment. The concept of complexity refers to many environmental variables beyond our control, comprehension, and prediction. Duffy suggests six successful DFs deal with a complex business environment 1.) flexibility and adaptability, 2.) ability to capture opportunities, 3.) intelligent management, 4.) integration of disruptive technologies, 5.) customer relationship, and 6.) learning organization. Overall, we rearrange these DFs into three elementary types of frames: a.) political & economic frames, b.) social and technological frames, and c.) ecological frames. Bismarck [105] said politics is the art of compromise. It is essential to be flexible and adaptable when framing the problem from a political perspective.

### 4.2.5.2 Organization Context

Like the environment frame (based-frame), the organizational frame has three elementary frames: organization design, rewarding systems, and information systems. An organization has its mission, goals, and objectives. Thus, organization design is to form a well-organized body to accomplish the defined goals and objectives. Organizational design is part of scientific management. One typical strategic organization design is a top-down approach that starts with the corporation's strategy implemented at headquarters and then cascades down to business groups, geographical units, and functional levels. Conversely, another way of organisational design is a bottom-up approach. It emphasizes the technological alignment between the social system and doing the work. Different strategies generate different organisational designs.

Nearly every organization starts with a single business strategy. It leads to a function-oriented organization design, including sales, marketing, operations, product development, finance and human resources. The organizational context is oriented by one vertical dimension because each functional head reports to the CEO. If the company seizes another business opportunity to develop a new business, it creates an independent unit that consists of an equal number of functions in the new business unit. One example is Amazon Web Services (AWS). The organisational context becomes two-dimensional since both CEOs of new and old business units have similar functional units. If the company expands its business globally, the organisational context should have a third-dimension consideration – a geographical dimension. The main challenge of organization design for DF is how to balance power and authority across a three-dimensional architecture.

Besides organisational design, the company also needs an information system because the organizational decision cannot leave without knowing inventory levels, sales costs, market demand, product schedule, and supply chains. The information should reflect how well the company performs. The performance of each functional unit is aligned with the company's reward system. Galbraith [106] articulated the star model to give holistic thinking about organisational context. There are many other models, such as the McKinsey 7-S model. However, the basic principles are the same.

### 4.2.5.3  Individual Context

Galbraith's star model has five essential components: strategy, structure, people, reward, and processes. The model demonstrates how these elements work together interactively. Apart from strategy, structure, reward, and processes, people or individuals play a critical role. With respect to people's context, Galbraith argued that instead of debating whether it should be dotted or solid lines of authority, it would add more value to an organization by focusing on the context of each individual's roles and responsibilities and then articulating various processes among them. Framing with an individual's context is to assign the right job to the right individual who has the right skillsets for the right responsibility. Practically, there are three practical guidelines 1.) knowing and searching for the right talent, 2.) matching the personality of the individual within the organizational culture, and 3.) developing people who understand the organization from multiple dimensions. The context frames represent relationships at three different levels: environment, organization, and individual. Apart from the relational context, we also need the material frames from a content perspective.

### 4.2.6  Content Frames

Framing from a content perspective is an inescapable part of the decision analysis process. [107] The word content has three implications: 1.) any material (e.g., data, information, electronic medium), 2.) subject, ideas or story (e.g., action, replay, clip), 3.) the amount of substance. "Content frame" is how to see something that is held together. It also means studying and understanding the implications, contexts, objectives, and actions for the given data or information. We can formulate three categorical frames for a given problem: policy, problem-solving, and perceiving[4]. These frames correspond to Macmillan and Tampoe's [108] classification of strategic management elements: strategic content, strategic thinking, and strategic action.

---

[4] As a reference point, this classification is similar to the AI definition, which is "the study of computations that make it possible to perceive, reason, and act".

### 4.2.6.1  Policy

The policy content is value-oriented and can be further classified into three elements: strategic policy, tactical planning, and operational control (Refer to Figure 4) [109] based on the scope and the end goal of a decision. Notice that these elementary frames could also be related to the contexts of events, agents, temporal, spatial, resources, and consequences.

For a policy frame, Porter [110] provided three critical principles of strategic policy decisions: 1.) "It is the creation of a unique and valuable position, involving a different set of activities." 2.) "It requires you to make trade-offs in competing—choose what not to do." 3.)" it involves creating 'fit' among a company's activities." These principles summarize the contents of a strategic policy. A policy is neither a set of best practices nor for operational improvement. A strategic policy is to define a plan of action to achieve the final goal. It is a set of guidelines, rules, and procedures to govern an organization or an entity for its activities. It sets up an activity's boundary for what should do and what should not. A good policy should yield today's actions to tomorrow's best possible outcomes.

A grand strategy implies that the scheme is at a much higher level. The time to reach the final goal requires much longer time and more resources. In contrast, tactical planning stands for engagement and execution of details or a part of the grand strategy. The operational control frame represents an immediate response to daily activities of problem-solving.

### 4.2.6.2  Problem-Solving

The content of the problem-solving frame is a solution-oriented formula. From a learning perspective, Steel [111] argues that "a problem is considered to be a matter which is difficult to solve or settle, a doubtful case, or a complex task involving doubt and uncertainty" This definition highlights three attributes: "complexity and intransparency," "require a solution," and "involving doubt and uncertainty." Jonassen [112] also suggests that the problem has two critical attributes: an unknown entity and solving unknown for some values.

In the cases of pedagogy and curriculum, the problem-solving process is often designed as a hands-on exercise. An instructor generally has a standard solution to a designed problem. This approach can be seen as the extension of medieval university's scholastic method [113]. The foundation of the scholastic method is disputation, in which the master or instructor very carefully formulates a problem. The solution to the problem has a binary answer. Unfortunately, the real-world problem is not binary, and problem-solving requires continuously searching and adapting. It is not a one-off resolution. In 1997, Levinthal [114] introduced the metaphor of the fitness landscape for managerial problem-solving, which is also described by Kauffman's NK model [115]. (N and K are two parameters)

McKelvey [116] and Scott [117] extends Kauffman's biological idea to the business and organizational arena. They summarize various complex problems into the basic three categories: "Mt Fuji", "Rugged", and "Dancing floor" problem landscape. The simple Mt Fuji problem landscape implies that we can discover the single peak value for the problem landscape through various decision apparatuses (Refer to figure 4). The value can be either a maximum or a minimum. On the other hand, the rugged landscape problem has many local peaks (local optimized values) but only one global peak. The global peak could be very hard to find without the right approach of balancing (trade-off) between exploring and exploiting processes. By comparison with rugged landscapes, the peak value of the dancing landscapes is always changing over time.

The traditional frame of problem-solving is not suitable for the fitness landscape. Klein and Klinger [118] proposed a naturalistic DM solution. The authors formulated ten different circumstances requiring "naturalistic decision-making". These are 1.) Ill-defined goals and ill-structured tasks, 2.) Uncertainty, ambiguity, and missing data, 3.) Shifting and competing goals, 4.) Dynamic and continually changing conditions, 5.) Action -feedback loops (real-time reactions to changed conditions), 6.) Time stress, 7.) High stakes, 8.) Multiple players, 9.) Organizational goals and norms, 10.) Experienced decision-makers. Klein and Klinger offered the recognition-primed decision (RPD) model to cope with various complex problems. Thus, the RPD gives rise to how we perceive a problem.

### 4.2.6.3   Perceiving

Hauser and Salinas [119] define the perceiving frame as "the process by which sensory information is used to guide behaviour toward the external world. This involves gathering information through the senses, evaluating and integrating it according to the current goals and internal state of the subject, and producing motor responses. In contrast to choice behaviour and decision making in general," "perceiving or perceptual decision making emphasizes the role of sensory information in directing behaviour (e.g., during a choice). Thus, within neuroscience, the goal is to reveal the computational mechanisms whereby neural circuits encode, store, and analyze perceptual signals; combine them with other behaviorally relevant information, and use them to resolve conflicts between competing motor plans." This definition emphasizes the perceptual signal inputs and computational thinking and analysis. It suggests DM is an interactive process between external perception signals and internal knowledge or experiences. Goldstein [120] highlights this process in Figure 6 with a series of nine steps that include three stimulus steps (1-3), three cognitive steps (4-6), and three reason-based steps (7-9).

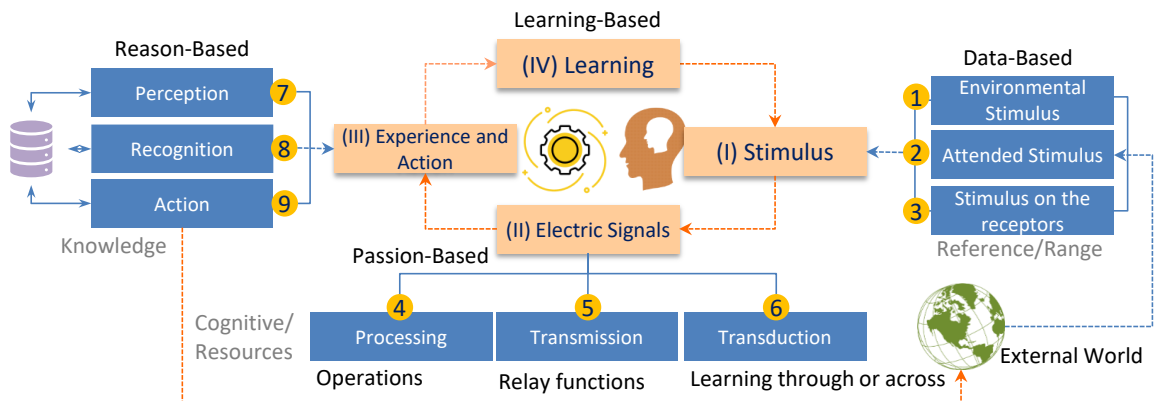

Figure 6: Personal Decision Making in a Simple Organism

These nine steps are roughly corresponding to four cognitive DM activities: (I.) stimulus, (II) electricity, (III) experiment & action, and (IV) learning. These activities further articulate three elementary frames 1.) Reason-based, 2.) Data-based (reference), and 3.) Passion-Based (cognitive resources/replacement).

## 5 IMPLICATIONS OF THE NEW TAXONOMY

### 5.1 Summary of Survey Results

Our survey illustrates that many researchers have provided various knowledge representation models for different applications since the 1980s. Some excellent research has [20] [30] laid the foundation for the DM paradigm. Others focused on a particular application [12] [121] [122] [123] [124] such as DM for the robotic navigation system [42] [43]. Still, others [13] [46] [47] [125] [126] [127] targeted some generalized issues for the particular domain of knowledge, such as problem-based learning, ethics, decision support systems, and managerial decisions.

The traditional way of the models is oriented by "rules plus database" (known as the GOFAI approach). The final output is to pursue or understand the decision itself. By contrast, this new taxonomy sets up a group of DFs as a searching landscape, in which we give a machine the desired output and database for a set of strategic rules. That is why we focus on the DF rather than a particular strategic decision.

The uniqueness of this comprehensive taxonomy is that it offers different types of knowledge representation models, such as emotion and value frames, through an interdisciplinary approach. This study aims at building a foundational framework from an AI perspective. The motivation is to leverage the AI/ML process for SD rules. In other words, we want to create the overall landscape of the SD problem. This landscape view is often considered as people's intuition or gut feeling. Clausewitz called it "passion".[19] Damasio [21] and Minsky [23] called it an "emotion machine". Clausewitz once argued that "Theory cannot equip the mind with formulas for solving problems, ... But it can give the mind insight into the great mass of phenomena and of their relationship."[19]. Based on the Newtonian methodology (iterative process for optimization), Clausewitz derives one of the most creative tools for Grand Strategy: the paradoxical trinity: Passion, Probability, and Reasons. We can derive the SDM learning trinity (Refer to Figure 7) from the paradox of trinity. It consists of irrationality (gut feeling, intuition, and emotions directed by illogic), non-rationality (data and probability driven by inductive logic) and rationality (knowledge and intelligence guided by deductive logic).

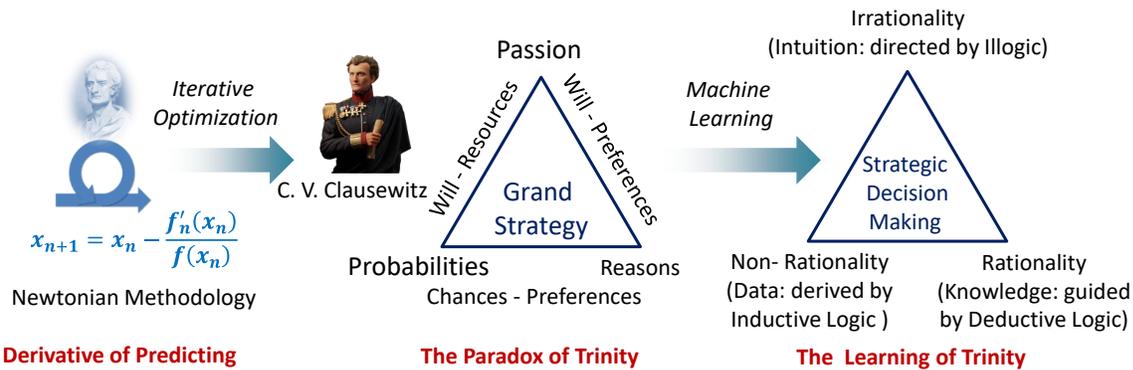

Figure 7: The Paradox of Trinity and Trinity of Learning

People's intuition or gut feeling is equivalent to Clausewitz's passion or Damasio and Minsky's emotions. Human emotions are an inescapable part of the DM process. Damasio [21] argues that emotions are in the loop of reason, and they could assist the reasoning process rather than necessarily disturb it. Without emotions, we could be in a forever deliberation loop. To better understand the role of emotions in the DM process, we could use the genetic algorithm as an analogy. In other words, human emotion is similar to the performance of the convergence rate ($\Delta\delta$) or stopping criteria in a genetic algorithm [128]. The new taxonomy with various DFs can help an SD maker to understand a problem landscape. Figure 8 illustrates the relationship between intuition or emotions, DFs, and a decision problem landscape. As we should see, a problem landscape might have one global maximum and a few local maximums. Some experienced people with good intuition can often call the shot (make up their mind) to stop further searching for the optimal global solution.

The study provides a comprehensive decision framework with 54 elementary frames. This is by no means exhaustive. We can generate more frames for a given strategic problem. However, we should also be aware that we have limited computational resources. The scope of DFs cannot be either too large or too small. We must balance the computational resources with a given strategic problem.

The fundamental issue is how to create a learnable framework for the machine to learn the problem landscape. A solution is to establish hierarchical or meta-learning or meta-meta-analysis models. It is the primary reason we build this new taxonomy in a three-layered architecture. Intuitively, we can select a set of combination strategic DFs, assign different parameters to each frame with a database and the desired outcome, and then let the machine tell us which set of DFs is most significant. This theme is beyond the scope of this paper.

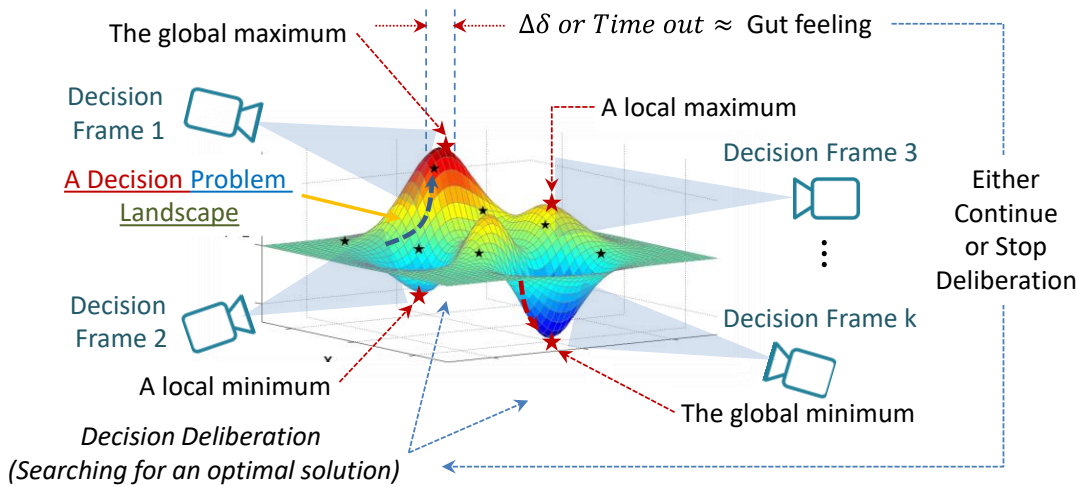

Figure 8: The Relationship between Gut Feeling, Decision Frames, and Decision Problem Landscape

In contrast to many previous studies [122] [124] [129], this study adopts a proper method for creating DFs taxonomy. We argue that although some previous models may still be helpful, they are not oriented from the AI perspective because many contemporary DFs' contexts are pretty subtle. Consequently, these models are unsuitable for uncertain, ambiguous, and chaotic environments.

## 5.2 The Characteristics of The New Taxonomy

Conversely, this taxonomy has seven distinguishing characteristics compared with many previous models: 1.) We focus on the theme of decision framing rather than SDM itself. By targeting the DF, we have the flexibility to cover a broad scope of decision models rather than focus on a particular application. It is particularly beneficial in addressing an SD problem, 2.) This taxonomy is an extension of Schoemaker's knowledge spectrum from an AI perspective, 3.) It is flexible. We should be able to apply these basic categories to a particular application, 4.)This decision framework has two dimensions: the knowledge spectrum (from certainty to chaos) and the logic reduction dimension, 5.) The new taxonomy allows us to leverage AI/ML searching for an optimal solution, 6.) Compared with previous taxonomies, this taxonomy included many mental frames, such as belief, mindset, culture, ethics, emotions etc. It allows us to tackle issues beyond traditional DFs. 7.) The uniqueness of this taxonomy is that it has multilevel abstraction. We can either add or subtract any number of elementary DFs. The architecture of the taxonomy is very elastic.

## 6 CHALLENGES, CONCLUSION, AND FUTURE DIRECTION

### 6.1 Challenges

One of the primary challenges in an SDM process is how to measure these DFs because of the value and fact gap. As Peter Drucker says, "you cannot manage what you cannot measure." We summarize future challenges as follows: 1.) How do we measure different types of DFs in terms of metrics (e.g., unit of emotions or feelings)? 2.) How do we construct different types of DFs? What is the methodology? 3.) How do we reconcile different parameters and allow a machine to search for possible rules on our behalf? 4.) How do we build various DFs searching algorithms for an optimal solution? 5.) Even if we could find an optimal solution, how can we explain and trust the result? 6.) How can we effectively control the learning trinity (irrational, non-rational, and rational frames)? 7.) How can we trade off some competing or even paradoxical values for different DFs?

### 6.2 Conclusion and Future Direction

We aim to develop a novel framework that enables us to adopt an AI/ML approach to reverse AI/ML programming logic and discover the optimal solution. We combine Bloom's taxonomy method, reductionism reasoning, and Schoemaker's knowledge spectrum to achieve such a goal. We develop a novel taxonomy that consists of 6 base frames, 18 categorical frames, and 54 elementary frames from a top-down perspective. From a bottom-up direction, DFs have emerged at multilevel abstraction.

The essence of this work is the art of possibility. We fill the knowledge gap in the taxonomy of DFs for SDM. It is the first time creating such a taxonomy to the best of our knowledge. Future directions are to consolidate this novel taxonomy, define the measurement unit for each frame, and test this representation model with various AI/ML algorithms for different SDM problems.

### ACKNOWLEDGMENTS

This research was funded in whole, or part, by the Luxembourg National Research Fund (FNR), grant reference C21/IS/16221483/CBD. For open access, the author has applied a Creative Commons Attribution 4.0 International (CC BY 4.0) license to any Author Accepted Manuscript version arising from this submission."

## A APPENDICES

Table 1: Various Decision Frames Proposed by Previous Researchers and Scholars

| Authors/ Method | Primary Contributions | Advantages | Potential Gaps |
|---|---|---|---|
| Carter et al. [32] / use qualitative cluster analysis and the Q-sort method | Highlight and categorize decision-making biases impact on supply chain management. They adopted the combination of qualitative cluster analysis and Q-sort methodology | They defined 76 decision biases and used qualitative data analyses to categorise them into 9 clusters. These decision biases can be generalized for decision frames. | However, this taxonomy is only applied to behavioural supply management practice rather than general decision-making. It focused on the decision biases. The proposed taxonomy required further testing |
| Paul C Nutt [24] / adopts both interviews and questionnaires method to collect data. | The author unveiled different factors of Decision Framing to influence the strategic direction from stakeholders' perspectives. It also offered an analytic view of SDM across different corporations. | A summary for seven types of SD from 317 firms shows only 61% of SD was made by executives & 39% by middle management. Framing shortfalls trigger a more successful decision. Framing conflicts, innovation & adaptation, lead to less successful decision | The analytic or quantified method is challenging to measure the success of SDs due to the nature of self-serving evaluation. Although some measurements of the SD can be quantified by rating, many of them are arbitrary. They can be opened for interpretation |
| Kris De Jaegher [33] / The paper employs prospect theory or method to model patients' frames of strategic decisions. | The study establishes a prospect-theoretic preferences model for health economics. Although it is behaviour-based modelling for strategic framing, the fundamental idea is more like game theory or a behavioural game | The research offers a theoretical model for a practical decision problem. According to Kahneman and Tversky's prospect theory, the author established a behaviour-based decision model for patients with different utility preferences (risk-averse or risk-taking). | It only provides the theoretical model for strategic framing between patients and physicians. There are no practical data and empirical analyses to support the theoretical model. The practical decision is much more complicated than a simple game theory. The model fails to consider the ethical frame |
| Richard J. Arend [34] / uses a logical approach with a three-step process for strategic decision-making under ambiguity (SDMUA). It adopts ex-post backward inductive logic reasoning with conditional probability | The paper defines a strategic decision problem with different choices over investment capital and payoff in ambiguous and uncertain environments. The authors extend from Decision-Making Under Ambiguity (DMUA) to SDMUA. They offer a simple three-step solution to cope with ambiguity. | The solution is to combine both game theory and incentive contract design. It defines SDMUA and theorizes the second-best approach as a new combination of economic tools. It offers a set of new prescriptions and predictions. The proposed approach identifies nine characteristics to deal with ambiguity from a managerial perspective. | The underlying structure is based on the combination of game theory with the expected utility payoff model. The model failed to provide experiment results to compare the proposed approach with intuition, pattern-following, history-based-extrapolation imitation, and stalling experimentation. Moreover, it does not show how to establish a cohesive system with these nine characteristics. |
| Cengiz Haksever et al.[35] / they apply value creation from profit, non-profit organizations' perspectives plus | The author defines the value creation proposition that includes both market (i.e. price) and nonmarket values (i.e. prestige, safety, and reliability) from the | The authors identify five possible scenarios for the impact of strategic decisions. 1.) create value for certain groups, not harm others, 2.) create value for one group, but harm others, 3.) | The paper fails to address other scenarios, such as creating value for all and destroying a part of stakeholders' value at another time…, etc. If we introduce a temporary domain, we can have at |
| temporary domain. It provide**s** the best practice **for** management | stakeholders' perspective. How to use this strategic decision model to improve key stakeholder relationship | destroy values for one group with no positive effects on others, 4.) destroy value for all, 5.) create values for all | least ten different scenarios that the author did not explore. |
| Paul Schoemaker [37] / He adopts Integrating organizational networks, weak signals, strategic radar (SR) & scenario planning | The authors provide methods how for leveraging SR or decision frames to integrate weak and emerging signals of threats or opportunities with sense-making, strategic dialogue and scenario planning | The authors offer five stages (setup, research, monitor, analyse, and publish) of SR or frame design or scenario-based system integrated with the scanning and monitoring external signals based on three assumptions. | The kind of SR is dependent on the leadership's intuition. It is not automatic. Unfortunately, the paper argued that most companies lack the ability to mine their extended network and to mind the weak signals in strategic ways (strategic framing). |
| Henry Mintzberg et al. [38]/ from multiple perspectives to define a concept of strategy | Different ways of defining the concept of strategy help both researchers and practitioners to understand how to manoeuvre through this difficult field. | Mintzberg et al. provide five practical ways of strategic framing or 5Ps from a planning perspective: Planning, Pattern, Position, Perspective, and Ploy | Further clarifying the meaning of strategy with the 5 Ps is not to explain how to implement a strategy but to remove some confusion. 5Ps do not offer computable algorithms or implementable actions. |
| Henry Mintzberg and Joseph Lampel [39]/ reflect on the strategy process / | The authors offer ten schools of thought for strategic management. Each school of thought has its philosophical root, tools, methods and concepts. | These frames are produced by ten schools of thought that can be divided into two groups: prescriptive (3 schools) = looking forward and descriptive (7 schools) = reasoning backwards | Each school of thought itself can be considered a kind of black box because the process of formulating strategy is not very clear. The issue with each school of thought for strategic framing is incomplete. |
| Thomas S Bateman & Carl P. Zeithaml [40] / Authors adopt the empirical method and found the strategic decision emerged from a stream of incremental decisions | Authors build the overall view of the strategic framing process as a series of incremental decisions. The study tested two hypothesises of the SD model: 1.) SD is an incremental process. 2.) SD is shaped by various contextual influences arising from the past and present and anticipates the future. | The model introduces the psychological content of the SDM process, which includes three decision stages; prior, the psychological context of the prospects of future gains and losses. It provides the empirical or ANOVA test results. The psychological context includes past, present, and future outlooks as three experimental variables with a feedback loop | It was built upon six assumptions, including 1.) failure feedback from the past experiences, 2.) significant higher levels of the repeated decisions, 3.) Decision-frames focus on gains rather than losses based on prospect theory. However, not all firms fit with this assumption, especially SMEs. The ANOVA result only shows an association relationship. The empirical data were collected from 193 students. |
| Marvin Minsky [23] / uses thought Experiments | The paper, together with his later book: "The Emotion Machine", can "open up the idea of possibilities that otherwise might be ignored or underestimated." | Minsky argued that human emotions are different ways of thinking for different problem types. He deliberately vague between computer science and psychology for next-generation research to fill their imagination. | Neuroscientists ask, "where are the data?" Philosophers ask, " Where are the proofs" Cognitive psychologists ask, where are the models that make testable predictions? Computer scientists ask, "Where is the code?" **[130]** |
| Marvin Minsky [30]/ An integrated methodology with the thought experiment | As a part of learning theory, The proposed frame theory to develop AI machines that would display human-like abilities. The concept is | Minsky's frame constructs as a way to represent knowledge in the machine. Frame-based knowledge representation can interact with the real world for | Minsky's hypothesis is an excellent thought experiment from an AI perspective. Still, many critics argue that there is no vigorous logical proof, no computer codes to be |
| | similar to Psychologist Bartlett's Schema theory and Philosopher Kuhn's paradigm theory. | new information. The frame is a data structure for representing a stereotyped situation | implemented for verification, and no data to support the frame hypothesis. |
| Bouton, Maxime [42] / He adopts Partially Observable Markov Decision Process (POMDP) with Simulation of Urban Mobility (SUMO) | Authors frame the problem of autonomous vehicles navigating urban intersections as a POMDP solution for coping with multiple conflict objectives. They provide empirical results and demonstrate that the POMDP is better than a threshold-based heuristic strategy regarding vehicles' s safety and efficiency. | It develops an online algorithm to cross an urban intersection autonomously. The POMDP is dependent on sampling from a generative model. The auto-system can dynamically change its decision to adapt to the behaviour of other agents. To track the vehicles, it adopts a multiple interacting model (IMM) filter | The solution does not consider sensor limitations (i.e. blockage). The results only compare with a simple heuristic or baseline policy, Time To Collision (TTC) threshold. It would be beneficial to compare online with offline algorithms and adopt a more accurate generative model than the linear Gaussian model. The result will be convinced if it includes pedestrians. |
| Christian Siagian et al. [43]/ deploy a hybrid topological/ grid-occupancy map that integrates the output from all perceptual models | It creates a new hybrid topological /grid-occupancy map that integrates the outputs from all perceptual modules of truly autonomous robots with the hierarchical representation framing. It is the first time to test a fully autonomous, visually guided localization and navigation system in a pedestrian environment | It creates a road recognition navigational system for both outdoor & indoor environments. It avoids both dynamic and static obstacles. It adopts monocular vision, complemented by Laser Range Finder (LRF) based obstacle avoidance. The perceptual modules include visual attention, landmark recognition, gist classification, location and road find. | Compared with the LRF-based system, it is very high. The success rate of the mission is dependent on the travel distance. The initial localization system depends on a pre-given database to guarantee correct convergence. The robot has not been tested in open areas and higher-density crowds. The system can not actively search for better views. |
| Christopher M. Scherpereel [46]/ Decision order Taxonomy. Method: Content analysis or semantic descriptors | Three Orders of Decision: 1st order of rational decision with deductive logic solutions. 2nd order of probabilistic uncertainty with inductive logic solutions. 3rd order of uncertainty, complex and dynamic with abductive logic and heuristic solutions | The author proposed an overarching terminology of a decision taxonomy. Overall, the author provided three layers of decision solutions or lenses for any given problem. It provided guidelines and a flowchart for many practitioners | The author did not address the emotional impact on the decision-making process. Not all problems can fit into these three categories. Also, many behaviour-based decisions depend more on pattern recognition rather than logical thinking. |

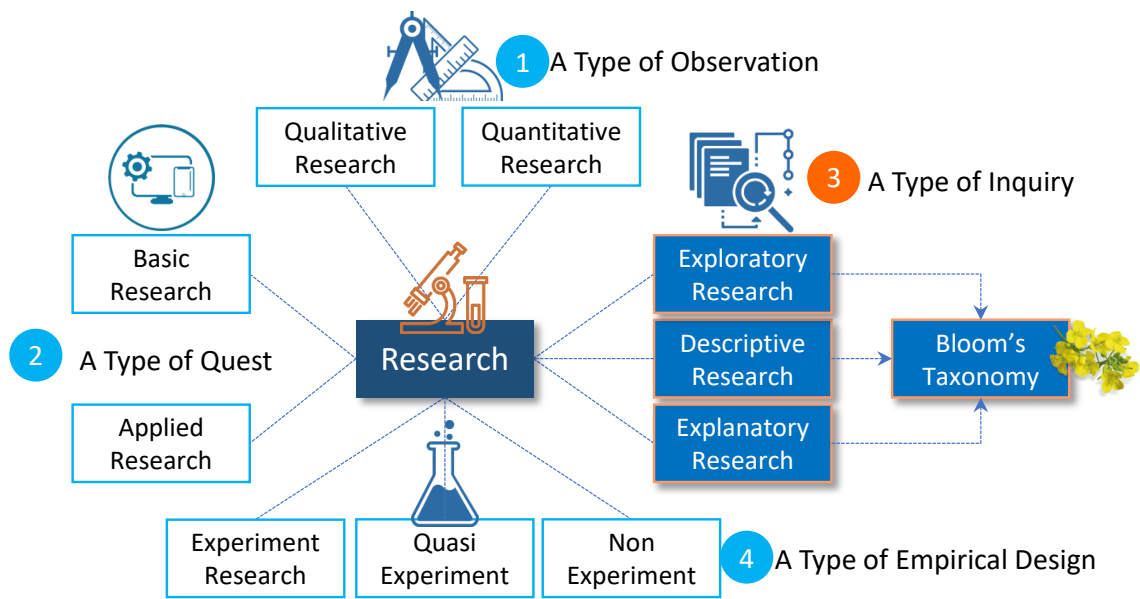

Figure 9: Four Categories of Research Methods